\documentclass[conference]{IEEEtran}
\IEEEoverridecommandlockouts
\usepackage[numbers,square,comma,sort&compress]{natbib}
\usepackage{url}
\usepackage[breaklinks,citecolor=blue,urlcolor=blue,bookmarks=false,hypertexnames=true]{hyperref}
\usepackage{amsmath,amssymb,amsfonts}
\usepackage{algorithmic}
\usepackage{graphicx}
\usepackage{textcomp}
\usepackage{multirow}
\usepackage{booktabs}
\usepackage{xcolor}
\usepackage{ulem}

\usepackage{etoolbox}
\makeatletter
\patchcmd{\@makecaption}
  {\scshape}
  {}
  {}
  {}
\makeatletter
\patchcmd{\@makecaption}
  {\\}
  {.\ }
  {}
  {}
\makeatother
\def\tablename{Table}

\makeatletter
\usepackage{xspace}
\DeclareRobustCommand\onedot{\futurelet\@let@token\@onedot}
\def\@onedot{\ifx\@let@token.\else.\null\fi\xspace}
\def\ie{\textit{i.e}\onedot}

\def\BibTeX{{\rm B\kern-.05em{\sc i\kern-.025em b}\kern-.08em
    T\kern-.1667em\lower.7ex\hbox{E}\kern-.125emX}}
\begin{document}

\title{\LARGE \bf{Talk2Radar: Bridging Natural Language with 4D mmWave Radar for 3D Referring Expression Comprehension} \\

% \thanks{This work is partially supported by the AI University Research Centre, Jiangsu Province Engineering Research Centre of Data Science and Cognitive Computation at XJTLU and Suzhou Municipal Key Laboratory for Intelligent Virtual Engineering (SZS2022004).
% This work received financial support from Jiangsu Industrial Technology Research Institute (JITRI).} 
% This part needs to be commented out when leaving to arxiv.
% \thanks{This work has been submitted to the IEEE for possible publication. Copyright may be transferred without notice, after which this version may no longer be accessible.} % This is required to avoid copyright prob when releasing to arxiv.
\thanks{$^{*}$ R. Guan, R. Zhang, N. Ouyang and J. Liu contribute equally.}
\thanks{
$^{1}$ Institute of Deep Perception Technology, JITRI, Wuxi, China

$^{2}$ Department of EEE, University of Liverpool, Liverpool, UK

$^{3}$ School of ECS, University of Southampton, Southampton, UK

$^{4}$ Momoni AI, Gothenburg, Sweden

$^{5}$ SAT, Xi'an Jiaotong-Liverpool University, Suzhou, China

$^{6}$ Thrust of Artificial Intelligence, HKUST (GZ), Guangzhou, China

$^{7}$ Thrust of Intelligent Transportation, HKUST (GZ), Guangzhou, China
}
\thanks{$^{\dag}$ Corresponding author: yutaoyue@hkust-gz.edu.cn}
}

\author{\IEEEauthorblockN{Runwei Guan$^{1,2,5\ *}$,
Ruixiao Zhang$^{3\ *}$,
Ningwei Ouyang$^{1,2,5\ *}$,
Jianan Liu$^{4\ *}$,
Ka Lok Man$^{5}$,
Xiaohao Cai$^{3}$, \\
Ming Xu$^{2,5}$, 
Jeremy Smith$^{2}$,
Eng Gee Lim$^{5}$, \textit{Senior Member, IEEE},
Yutao Yue$^{6,7,1\ \dag}$,
Hui Xiong$^{6}$, \textit{Fellow, IEEE} 
}}
% \IEEEauthorblockA{\IEEEauthorrefmark{1}Institute of Deep Perception Technology, Jiangsu Industrial Technology Research Institute, Wuxi, China; \\
% Email: \{guanrunwei, ouyangningwei, yueyutao\}@idpt.org}
% \IEEEauthorblockA{\IEEEauthorrefmark{2}School of Advanced Technology, Xi'an Jiaotong-Liverpool Unversity, Suzhou, China;\\ 
% Email: \{runwei.guan, ningwei.ouyang\}@liverpool.ac.uk, \{ka.man, enggee.lim\}@xjtlu.edu.cn}
% \IEEEauthorblockA{\IEEEauthorrefmark{3}Department of Electronics and Computer Science, University of Southampton, Southampton, United Kingdom;\\ 
% Email: rz6u20@soton.ac.uk}
% \IEEEauthorblockA{\IEEEauthorrefmark{4}Vitalent Consulting AB, Gothenburg, Sweden; Email: jianan.liu@vitalent.se}
% \IEEEauthorblockA{\IEEEauthorrefmark{5}Department of Electrical Engineering and Electronics, University of Liverpool, Liverpool, United Kingdom; \\
% Email: \{runwei.guan, j.s.smith\}@liverpool.ac.uk}
% \IEEEauthorblockA{\IEEEauthorrefmark{6}Thrust of Artificial Intelligence and Thrust of Intelligent Transportation, HKUST (Guangzhou), GuangZhou, China; \\
% Email: yutaoyue@hkust-gz.edu.cn}
% }

\maketitle

\begin{abstract}
    Embodied perception is essential for intelligent vehicles and robots in interactive environmental understanding. However, these advancements primarily focus on vision, with limited attention given to using 3D modeling sensors, restricting a comprehensive understanding of objects in response to prompts containing qualitative and quantitative queries. Recently, as a promising automotive sensor with affordable cost, 4D millimeter-wave radars provide denser point clouds than conventional radars and perceive both semantic and physical characteristics of objects, thereby enhancing the reliability of perception systems. To foster the development of natural language-driven context understanding in radar scenes for 3D visual grounding, we construct the first dataset, Talk2Radar, which bridges these two modalities for 3D Referring Expression Comprehension (REC). Talk2Radar contains 8,682 referring prompt samples with 20,558 referred objects. Moreover, we propose a novel model, T-RadarNet, for 3D REC on point clouds, achieving State-Of-The-Art (SOTA) performance on the Talk2Radar dataset compared to counterparts. Deformable-FPN and Gated Graph Fusion are meticulously designed for efficient point cloud feature modeling and cross-modal fusion between radar and text features, respectively. Comprehensive experiments provide deep insights into radar-based 3D REC. We release our project at \textcolor{magenta}{\url{https://github.com/GuanRunwei/Talk2Radar}}.

 % Embodied perception is essential for intelligent vehicles and robots, enabling more natural environmental understanding and interaction. However, these advancements currently embrace vision, rarely focusing on using 3D modeling sensors, which limits the full understanding of surrounding objects with multi-granular characteristics. Recently, as a promising automotive sensor with affordable cost, 4D millimeter-wave radars provide denser point clouds than conventional radars and perceive both semantic and physical characteristics of objects, thus enhancing the reliability of perception system. To foster the development of natural language-driven context understanding in radar scenes for 3D grounding, we construct the first dataset, Talk2Radar, which bridges these two modalities for 3D Referring Expression Comprehension (REC). Talk2Radar contains 8,682 referring prompt samples with 20,558 referred objects. Moreover, we propose a novel model, T-RadarNet for 3D REC upon point clouds, achieving state-of-the-art performances on Talk2Radar dataset compared with counterparts, where deformable-FPN and gated graph fusion are meticulously designed for efficient point cloud feature modeling and cross-modal fusion between radar and text features, respectively. Further, comprehensive experiments are conducted to give a deep insight into radar-based 3D REC. We release our project at \textcolor{magenta}{\url{https://github.com/GuanRunwei/Talk2Radar}}.
\end{abstract}

% \begin{IEEEkeywords}
% 3D referring expression comprehension, 3D visual grounding, 4D mmWave radar, multi-modal fusion
% \end{IEEEkeywords}

% \begin{figure}[htbp]
% \begin{center}
% \includegraphics[width=0.92\linewidth]{figures/overview.pdf}
% \end{center}
% \vspace{-3mm}
% \caption{Overview of proposed pipeline in Talk2Radar for 3D REC.}
% \label{fig:overview}
% \end{figure}

\section{Introduction}
Millimeter-wave (mmWave) radar, as an all-weather and low-cost perception sensor \cite{yao2023radar1,yao2023radar2} which can capture objects' distance, azimuth, velocity, motion directions, reflected power, etc., has been widely used in perception for autonomous driving \cite{deep_radar_seg, zeller2023radar, popov2023nvradarnet,RaLiBEV,RCM-Fusion,CRN}, robotic navigation \cite{radar_SLAM,radar_SLAM_2} and Cooperative Intelligent Transportation Systems (C-ITS) \cite{V2X_CP_Overview}. Recently, to enhance detection accuracy, 4D radar emerges and addresses the limitation of conventional radars' inability to measure height and significantly increases Point Cloud (PC) density \cite{4d_radar_survey,4d_radar_survey_TIV}, allowing objects to contain more exploitable features for fruitful scene understanding and better performance in downstream tasks on land \cite{RCFusion, LXL, 4D_Radar_SLAM, 4D_Radar_SLAM_2, 4D_Radar_MOT, RCBEVDet, DPFT, paek2022k} and maritime \cite{guan2023achelous,guan2024mask,yao2023waterscenes,guan2024watervg,Efficient_vrne}.

Meanwhile, recent advancements driven by Vision-Language Models (VLMs) \cite{wang2023drive} in embodied intelligence and human-centric intelligent driving perception systems \cite{cui2024survey} enable intelligent vehicles and robots to understand human commands, perceive surroundings, and make decisions \cite{xie2024large}. However, these advancements are primarily confined to the visual domain \cite{sima2023drivelm, liu2023gres, wu2023referring}. While some research has focused on 3D visual grounding, these efforts are predominantly aimed at RGB-D and LiDAR \cite{achlioptas2020referit3d, zhao20213dvg, cheng2023language, yang2023lidar, hess2024lidarclip}. For autonomous driving and navigation, we argue that referring prompts should not only emphasize positional relations, shapes, and categories but also include object motion characteristics, which is precisely the strength of radar. Radar is immune to adverse weather conditions and can provide quantified measurements of distance, velocity, motion, and azimuth, while Radar Cross Section (RCS) and Point Clouds (PCs) offer qualitative semantic features. This dual capability allows for a comprehensive description of objects, incorporating both qualitative and quantitative characteristics. With the emergence of high-resolution 4D radar, this unexplored field is poised to be addressed. Unlike VLM-based environmental understanding, which can leverage web-scale training data, radar data depend on real-world scenarios and physical models \cite{pushkareva2023radar}. Currently, there are no relevant datasets or research focusing on this area. Hence, we aim to pioneer and unlock the potential of 4D radar for natural language-based multi-modal 3D object localization. Specifically, this enables individuals to locate specific objects using 3D bounding boxes by describing features perceivable by 4D radar.

\begin{figure*}
\begin{center}
\includegraphics[width=0.99\linewidth]{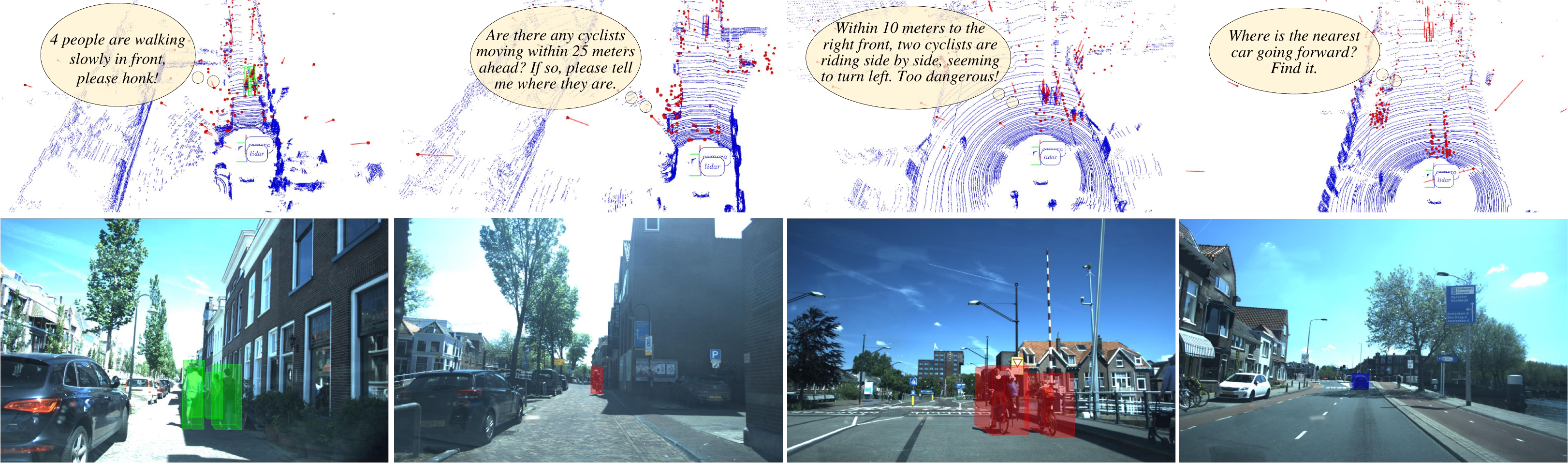}
\end{center}
\vspace{-6mm}
\caption{Samples in Talk2Radar. The first and second row respectively presents referred objects by point cloud sensors (\textcolor{red}{4D radar} and \textcolor{blue}{LiDAR}) and camera.}
\label{fig:dataset_samples}
\end{figure*}

Based on these foundations, we introduce 3D Referring Expression Comprehension (REC), also known as open-vocabulary 3D object detection or 3D visual grounding \cite{10531702}, which locates specific objects based on textual descriptions. To this end, we establish the first radar-based 3D visual grounding dataset, Talk2Radar, and its corresponding benchmark. Talk2Radar is built upon the well-known 4D radar dataset, View of Delft (VoD) \cite{palffy2022multi}. Each sample in Talk2Radar contains radar PCs, LiDAR PCs, an RGB image, and a textual reference prompt. The dataset possesses the following two key characteristics: \textbf{(i)} A prompt can refer to one or more objects, making it more flexible and realistic; \textbf{(ii)} The text features contain object attributes perceivable by radar, while excluding any features perceivable by vision. Besides the dataset, we propose a novel radar-text fusion model called T-RadarNet for 3D visual grounding. Within T-RadarNet, \textbf{firstly}, we introduce Deformable-FPN to enhance the modeling of irregular radar PCs. \textbf{Secondly}, to efficiently fuse the features of radar and text, we devise a module named Gated Graph Fusion (GGF). GGF leverages graph and gating mechanisms for neighborhood feature association and cross-modal feature matching. Notably, GGF can be integrated into most PC detectors for 3D visual grounding, demonstrating impressive generalization capabilities.

The contributions of this paper are summarized as follows:

\begin{enumerate}
\item We present Talk2Radar, the first 4D radar-based 3D REC dataset for terrestrial autonomous driving, along with its corresponding benchmark for subsequent studies.
\item For 4D radar-based 3D visual grounding, we propose a novel 3D REC model named T-RadarNet, by incorporating the devised Deformable-FPN and GGF.
\item Comprehensive experiments have been conducted to analyze and enhance 4D radar-based 3D visual grounding, to promote a thorough understanding in this field.
% \item An open-source 3D visual grounding framework tailored for point cloud sensors under traffic scenarios.
\end{enumerate}
\vspace{-2mm}

\section{Related Works}

\subsection{3D Referring Expression Comprehension in Traffic}
As a challenging task based on multi-modal learning and 3D geometry, 3D REC aims to provide a textual prompt and locate one or more objects most closely matching it in the form of 3D bounding boxes. \textbf{(i) Datasets.} As Table. \ref{tab:dataset_compare} shows,  Talk2Car \cite{deruyttere2019talk2car} is a 2D image-based REC dataset on driving scenarios upon Nuscenes \cite{caesar2020nuscenes}, but providing LiDAR data of the same frame. M3DRef \cite{zhan2024mono3dvg} is a monocular 3D grounding dataset upon KITTI \cite{geiger2012we}. NuPro \cite{wu2023language} is a 3D grounding benchmark based on multi-camera and multi-frames for autonomous driving. \textbf{(ii) Methods.} \cite{zhan2024mono3dvg} proposes a monocular-based 3D REC baseline model called Mono3DVG-TR. \cite{solgi2024transformer} raises a model called AFMNet for modeling complex 3D object relation. \cite{cheng2023language} propose a baseline model for LiDAR-based visual grounding upon Talk2Car dataset. Nevertheless, radar-based 3D REC is still unexplored and lacks full-scale analysis. Moreover, textual prompts of our Talk2Radar dataset contain more attributes of objects and context with qualitative or numerical description, and provide a realistic referring paradigm.

% \cite{chen2020scanrefer} propose a 3D indoor REC dataset called ScanRefer upon RGB-D scans. \cite{achlioptas2020referit3d} construct a 3D indoor REC dataset called ReferIt3D, where one prompt refers to single or more objects.
% \cite{roh2022languagerefer,huang2022multi,guo2023viewrefer,chen2022language} are 3D indoor visual grounding models. 

\subsection{3D Object Detection with Point Cloud}
3D object detection is vital for environmental perception for autonomous driving. PC sensors mainly include LiDAR and 4D radar. LiDAR can provide rich geometric and depth features \cite{liu2024difflow3d,liu2025dvlo}, while 4D radar can capture certain semantic features, motion, velocity, and depth features of objects \cite{jia2025radarnext,zheng2025doracamom,zhang2024radarode}. Currently, PC-based 3D detectors can be primarily divided into pillar-based and voxel-based. \cite{lang2019pointpillars,li2023pillarnext,mao2024pillarnest} are three fast pillar-based, which project PCs onto a Bird's-Eye View (BEV) and then extract features like image processing. \cite{zhou2018voxelnet, yan2018second, chen2023voxelnext} are three voxel-based detectors converting PCs into voxel grids, which are known for simplicity, but they may suffer from low computation efficiency. Additionally, \cite{yin2021center, zhou2022centerformer} provide an anchor-free paradigm for 3D detection. \cite{liu2023smurf} fuses different representations of PCs dedicated to radar-based 3D detection. Based on the above, we fully consider two aspects in our T-RadarNet: \textbf{(i)} Extraction of irregular PC features, \textbf{(ii)} Dynamic weighting and separation of objects and clutter for high-quality fusion with text features.

\begin{table}
\setlength\tabcolsep{3.5pt}
    \caption{Comparison of 3D visual grounding datasets in traffic}
    \vspace{-3mm}
    \label{tab:dataset_compare}
    \centering
    \begin{tabular}{l|c|ccc}
    \toprule
     \textbf{Datasets}  & \textbf{Sensors} & \textbf{Objects} &  \textbf{AvgExpr} & \textbf{Context} \\
    \hline
     % ScanRef \cite{chen2020scanrefer} & Indoor & RGBD Cam & 11,046 & 20.27 & depth size location \\
     % \midrule
     % RefIt3D \cite{achlioptas2020referit3d} & Indoor & RGBD Cam & 1,4742 & 11.40 & color size location\\
     % \midrule
     \multirow{2}[2]{*}{NuPro \cite{wu2023language}} & \multirow{2}[2]{*}{Camera} & \multirow{2}[2]{*}{187,445} & \multirow{2}[2]{*}{-} & color, size, relation  \\
      & & & & motion, location \\
    \hline
    \multirow{2}[2]{*}{Talk2Car \cite{deruyttere2019talk2car}} & Camera, & \multirow{2}[2]{*}{10,519} & \multirow{2}[2]{*}{11.01} & color, size, \\
     & LiDAR & & & location, relation \\
    \hline
    \multirow{2}[2]{*}{M3DRef \cite{zhan2024mono3dvg}} & \multirow{2}[2]{*}{Camera} & \multirow{2}[2]{*}{8,228} & \multirow{2}[2]{*}{53.24} & size, relation,\\
     & && & location, color \\
    \hline
    \multirow{2}[2]{*}{\textbf{Talk2Radar}} & \textbf{4D radar,} & \multirow{2}[2]{*}{\textbf{20,558}} & \multirow{2}[2]{*}{\textbf{14.30}} & \textbf{size, motion, location,} \\
     & \textbf{LiDAR} & & & \textbf{relation, velocity, depth} \\
    \bottomrule
    \end{tabular}
\end{table}
% \vspace{-1mm}

\section{Talk2Radar Dataset}

\begin{table*}
\setlength\tabcolsep{1.2pt}
    \centering
    \caption{Statistics of referent object number and point clouds in Talk2Radar}
    \label{tab:dataset_point_clouds}
    \vspace{-3mm}
    \begin{tabular}{c|cccccccccccc}
    
    \hline
       \textbf{Objects}  & \multirow{2}[2]{*}{\textbf{Pedestrian}} & \multirow{2}[2]{*}{\textbf{Cyclist}} & \multirow{2}[2]{*}{\textbf{Car}} & \multirow{2}[2]{*}{\textbf{motor}} & \multirow{2}[2]{*}{\textbf{truck}} & \multirow{2}[2]{*}{\textbf{bicycle}} &  \multirow{2}[2]{*}{\textbf{rider}} &  \textbf{moped} & \textbf{bicycle} & \textbf{human} & \textbf{rider} & \textbf{vehicle}  \\
         & & & & & & & & \textbf{scooter} & \textbf{rack} & \textbf{depiction} & \textbf{other} & \textbf{other} \\
    \hline
      \textbf{Sensor PC} & 3487 (17.0\%) & 3234 (15.8\%) & 9336 (45.6\%) & 73 (0.4\%) & 27 (0.1\%) & 2442 (11.9\%) & 159 (7.8\%) & 470 (2.3\%) & 1165 (5.7\%) & 48 (2.3\%) & 11 (-) & 3 (-)\\
    \hline
      Radar1$_{\text{(1-frame)}}$ & 2.0 & 3.8 & 3.6 & 3.4 & 17.7 & 1.4 & 1.7 & 1.8 & 3.0 & - & 2.5 & - \\
      Radar3$_{\text{(3-frame)}}$ & 5.3 & 9.6 & 11.8 & 19.1 & 65.2 & 4.3 & 3.6 & 7.1 & 9.8 & - & 8.0 & - \\
      Radar5$_{\text{(5-frame)}}$ & 7.4 & 12.1 & 18.7 & 44.4 & 128.0 & 7.0 & 4.2 & 9.6 & 14.3 & - & 9.5 & - \\
    \hline
      LiDAR & 258.9 & 211.5 & 771.6 & 1378.3 & 7338.4 & 174.2 & 273.9 & 232.9 & 215.5 & - & 6.0 & - \\
    
    \hline
    \end{tabular}
\end{table*}

\subsection{Data Collection and Annotation Process}
\begin{figure}
\begin{center}
\includegraphics[width=0.99\linewidth]{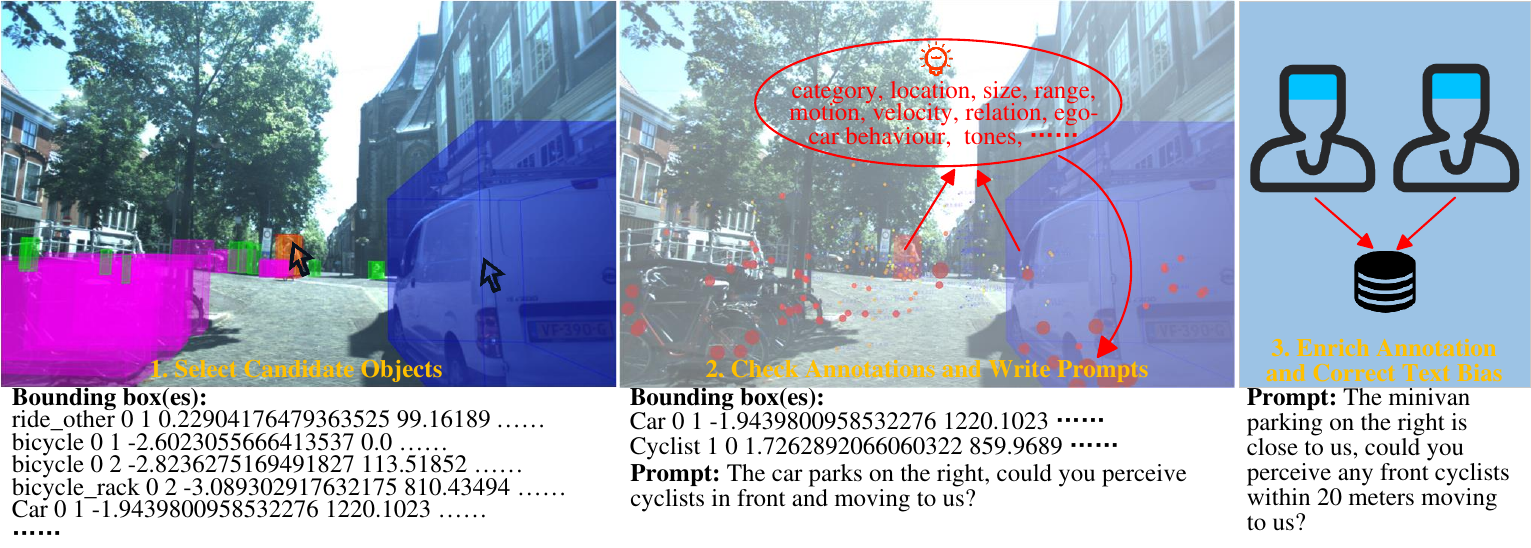}
\end{center}
\vspace{-6mm}
\caption{Annotation process of Talk2Radar dataset.}
\label{fig:annotation_process}
\end{figure}

Thanks to the renowned VoD dataset \cite{palffy2022multi} in the field of radar perception, which is specifically designed for autonomous driving perception and equipped with the ZF FRGen21 4D mmWave radar, as well as a LiDAR and a stereo camera, we can engineer textual prompts based on finely annotated 3D bounding boxes of various road objects. Fig. \ref{fig:dataset_samples} shows some samples, while Fig. \ref{fig:annotation_process} presents the entire annotation process.

\textbf{(1) Select Candidate Object(s):} First, utilizing the developed Graphical User Interface for annotation, we present the image with projected 3D bounding boxes for easy observation. Then, experienced annotators select the candidate boxes in the image that need to be described. The corresponding 3D bounding box annotation is saved. The annotator prioritizes selecting objects within the radar's Field of View (FoV).

\textbf{(2) Check Annotations and Write Prompts:} Based on the chosen 3D bounding box(es), the annotator first checks for any errors in the position or category of bounding boxes. During annotation, we also attach compensated vector velocity and plane depth next to the corresponding PC for reference. Considering detection errors and object size, we generally describe the object depth as a range rather than a fixed value. For individual objects, annotators consider attributes such as category, spatial position relative to the ego vehicle, distance, velocity, motion trend, and size, but exclude color and other information that radar cannot perceive. If multiple objects are selected, their spatial relationships are also considered.

\textbf{(3) Enrich Annotation and Correct Text Bias:} To avoid subjectivity and errors in text prompts, we implement a collaborative review and revision process involving two additional annotators after the initial annotation. This process serves to increase the diversity of descriptions to test the model's robustness and generalization. Additionally, it allows for the discussion of subjective errors made by the initial annotator, followed by corrections upon reaching a consensus.

\subsection{Dataset Statistics}
We present quantitative statistics from the perspectives of textual prompts and referent objects.

\begin{figure}
\begin{center}
\includegraphics[width=0.99\linewidth]{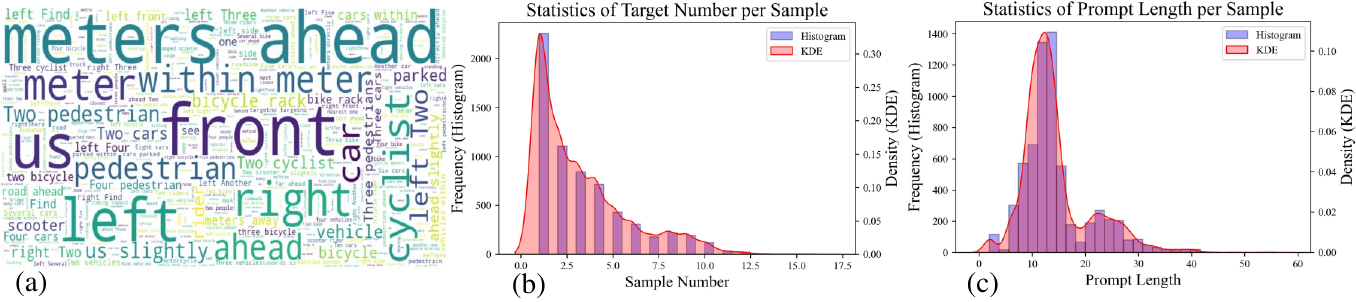}
\end{center}
\vspace{-6mm}
\caption{Statictics of Talk2Radar dataset on referent objects and prompts.}
\label{fig:dataset_statistics}
\end{figure}

\textbf{Textual Prompts:} As illustrated in Fig. \ref{fig:dataset_statistics}(a), \textbf{firstly}, high-frequency words in the word cloud include both qualitative and quantitative descriptions. This rich vocabulary benefits from the 4D radar's multi-dimensional perception capabilities, capturing quantitative features such as object velocity, motion direction, and depth, while also perceiving some degree of semantic features. This enables more detailed descriptions of specific objects and allows for fine-grained object clustering and filtering based on individual or partial object attributes. \textbf{Secondly}, as shown in Fig. \ref{fig:dataset_statistics}(c), the prompt length distribution in Talk2Radar is broad, with an average value of 14.30 (Table \ref{tab:dataset_compare}, \textbf{AvgExpr}). This variety allows the model to learn a wide range of sentence patterns, increasing task complexity as the model must handle text prompts of varying patterns, structures, and lengths, correctly understanding their semantic representations to query complex PC scenes.

\textbf{Referent Objects:} \textbf{Firstly}, as presented in Table \ref{tab:dataset_compare}, Talk2Radar contains a total of 20,558 referent objects. \textbf{Secondly}, Table \ref{tab:dataset_point_clouds} shows the distribution of object quantities across 12 categories and the average PC quantity for each category, including radar (1, 3, and 5 frames accumulated) and LiDAR. Cars, pedestrians, and cyclists are the three primary referent categories in Talk2Radar. Despite the improved resolution of 4D radar, there remains a gap in PC density compared to LiDAR. \textbf{Thirdly}, as shown in Fig. \ref{fig:dataset_statistics}(b), the number of referent objects per sample in Talk2Radar ranges from 1 to 11, presenting a challenge for understanding complex scenes. Specifically, the model must adaptively filter out noise words in prompts, focus on the key characteristics of the referent object(s), and perform multiple Region-Of-Interest (ROI) queries in irregular PC contexts.

\subsection{Metrics and Subset Settings}

\textbf{Metrics:} We adopt Average Precision (AP) and Average Orientation Similarity (AOS), including 3D bounding box mAP and mAOS on the entire annotated and driving corridor area, consistent with VoD for comprehensive evaluation.

\textbf{Subset Settings:} We divide Talk2Radar into three subsets, sharing same IDs with VoD. The number of subsets for training, validation and test are 5139, 1296 and 2247, respectively. Since the test set annotation of VoD is not publicly available, models are evaluated on the validation set in this paper.

\begin{figure*}
\begin{center}
\includegraphics[width=0.99\linewidth]{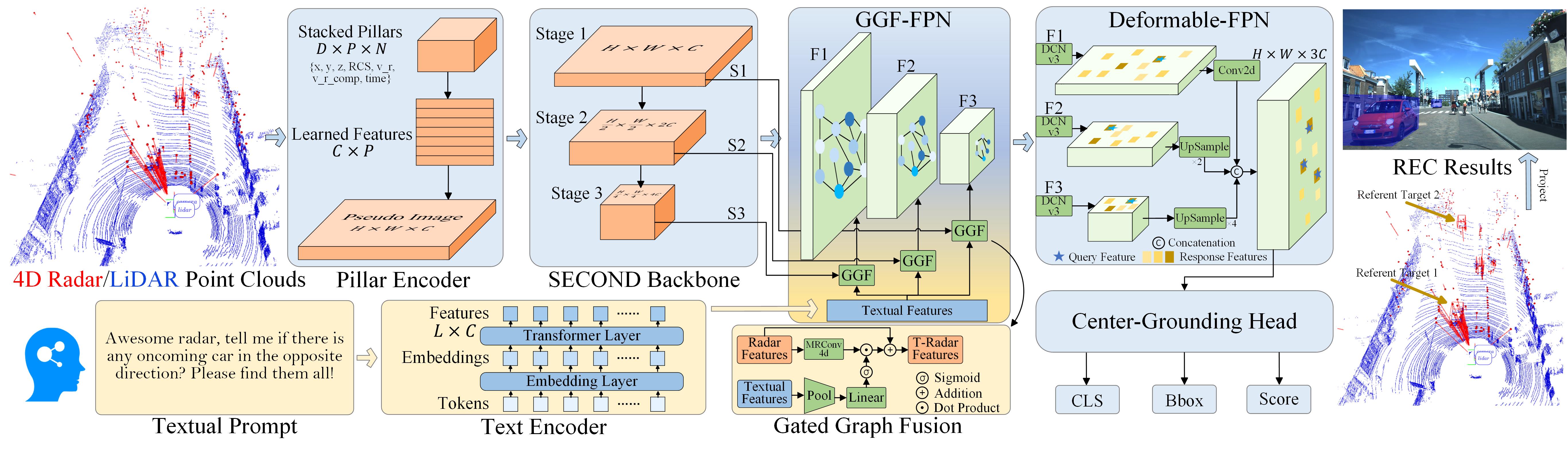}
\end{center}
\vspace{-6mm}
\caption{The architecture of \textbf{T-RadarNet}. LiDAR is not the main modality, but it can also be used as the input of T-RadarNet.}
\label{fig:t_radarnet}
\end{figure*}
\vspace{-2mm}

\section{Methods}

\subsection{Overall Pipeline}
Fig. \ref{fig:t_radarnet} illustrates the detailed architecture of T-RadarNet. Given a frame of radar PCs and a textual prompt as input, T-RadarNet provides the 3D grounding prediction of the specified object(s) guided by natural language. Firstly, for primary PC feature extraction, considering computational efficiency and scalability, we adopt a pillar encoder for the basic representation of 3D PC as a 2D pseudo image. Subsequently, the concise SECOND backbone \cite{yan2018second} is used to extract three-stage radar PC features \(\{F^{S_1}_R, F^{S_2}_R, F^{S_3}_R\}\) at multiple scales. Secondly, we use transformer-based encoders (e.g., ALBERT) \cite{wolf2019huggingface} for dynamic context representation of the textual prompt. Thirdly, given the interference in radar PC context from multi-path clutter, aligning and fusing dense textual semantic features with sparse PCs pose significant challenges. To address this, we design a graph-based strategy called Gated Graph Fusion (GGF), which associates ROI in the radar feature space and applies point-wise cross-modal gating between radar and text, yielding the most matched PC regions aligned with the text. The three-stage PC features and textual feature \(F_T \in \mathbb{R}^{L \times C}\) are then fed into GGF-FPN, which consists of GGF modules for cross-modal fusion, outputting three-scale text-conditional radar features \(\{F_{R|T}^{S_1}, F_{R|T}^{S_2}, F_{R|T}^{S_3}\}\). Fourthly, given the irregularity and sparsity of radar PC, conventional convolutions struggle to interact with them. Hence, we propose Deformable-FPN based on deformable convolutions \cite{wang2023internimage}, which provides high-quality PC features queried by the textual prompt to the grounding head for localization. Lastly, considering the efficiency of anchor-free detectors, we adopt the center-based detection head \cite{yin2021center} and employ the deformable convolution-based separable localization head for text-referenced 3D detection.

\subsection{Gated Graph Fusion}
For one-stage deep radar feature $F_R \in \mathbb{R}^{H \times W \times C_r}$, we first construct a graph map $\mathcal{G}=H(F_R)$ to aggregate neighbour features and associate potential object areas, where $F_R = \{F_R^1, F_R^2, \dots, F_R^i\}$ denotes a set of unordered nodes initialized upon \cite{li2019deepgcns}. We implement the graph convolution $H(\cdot)$ as: %Eq. (\ref{eq:gcn1}) presents:
\begin{equation}
    \begin{aligned}
    \mathcal{G} & = H(F_R, \mathcal{W}) \\
    & = {\rm Update}({\rm Aggregate}(F_R, W_{\rm agg}), W_{\rm update}),
    \end{aligned}
    \label{eq:gcn1}
\end{equation}
where $W_{\rm agg}$ and $W_{\rm update}$ are two learnable weights for feature aggregation and update. Specifically, the aggregation and update operations calculate the representation of the current node $i$ by aggregating features of neighbouring nodes, \ie,
% which is presented in Eq. (\ref{eq:gcn2}):

\begin{equation}
    \hat{F}_R^i = h(F_R^i, g(F_R^i, \mathcal{N}(F_R^i), W_{\rm agg}), W_{\rm update}),
    \label{eq:gcn2}
\end{equation}
where $\mathcal{N}(F_R^i)$ is the set of neighbour nodes of $F_R^i$. Considering the efficiency of graph feature aggregation, we first adopt Max-Relative Graph Convolution (MRConv4d) \cite{li2019deepgcns}. To enrich context, we utilize a residual path with concatenation $\oplus$ to construct radar graph with neighourhood aggregated features:
% which is shown 
% %in Eq. (\ref{eq:gcn3}) and (\ref{eq:gcn4}) as
% below:
\begin{align}
    & g(\cdot) = \tilde{F}_R^i = \max(\{F_R^i - F_R^j \ | \ j \in \mathcal{N}(F_R^i)\}) W_{\rm agg}, \label{eq:gcn3} \\
    & h(\cdot) = \hat{F}_R^i = \tilde{F}_R^i W_{\rm update} \oplus F_R^i, \label{eq:gcn4}
\end{align}

Based on the above, we construct the graph-based deep radar feature $F_{\mathcal{G}(R)} \in \mathbb{R}^{H \times W \times C_r} = \{\hat{F}_R^1, \hat{F}_R^2, \dots, \hat{F}_R^i\}$.

For the textual feature $F_T \in \mathbb{R}^{L \times C}$, we first adopt max-pooling along the token dimension to extract abstract high-level semantic information. Then, to align with the radar graph feature $F_{\mathcal{G}(R)}$, a linear transformation is exerted on it and we obtain the abstract textual feature $\hat{F}_{T} \in \mathbb{R}^{1\times 1 \times C}$. Further, to effectively suppress noise and focus on object ROI, we adopt a cross-modal gating strategy. Exactly, we obtain the weights of the linearly transformed textual features through Sigmoid activation and multiply them point-wise with the radar graph feature $F_{\mathcal{G}(R)}$. Finally, the feature is enhanced by addition with $F_{\mathcal{G}(R)}$ by a residual path. The process is shown below:
%in Eq. (\ref{eq:ggf_1}) and (\ref{eq:ggf_2}):
\begin{align}
     \hat{F}_{T} & = {\rm MaxPool}(F_T), \label{eq:ggf_1}  \\
     F_{R|T}  & = F_{\mathcal{G}(R)} \odot \sigma(\hat{F}_{T} \cdot W_T) + F_{\mathcal{G}(R)}, \label{eq:ggf_2} 
\end{align}
where $\odot$ and $\sigma$ denote the point-wise multiplication and Sigmoid function, respectively.

\begin{table*}
    \setlength\tabcolsep{3.4pt}
    \caption{Overall performances on Talk2Radar (\textcolor{red}{\textbf{Best Radar}}, \textcolor{blue}{\textbf{Best LiDAR}}, \uline{Best Performance}. \textbf{Car}, \textbf{Pedestrian} and \textbf{Cyclist} provide specialized \textbf{mAPs}.)}
    \vspace{-3mm}
    \label{tab:benchmark_compare}
    \centering
    \begin{tabular}{c|ccc|cccc|c|cccc|c}
    \hline
       \multirow{2}[2]{*}{\textbf{Models}} & \multirow{2}[2]{*}{\textbf{Sensors}} & \multirow{2}[2]{*}{\textbf{Text Encoder}} & \multirow{2}[2]{*}{\textbf{Fusion}} & \multicolumn{5}{c}{\textbf{Entire Annotated Area (EAA)}}  & \multicolumn{5}{c}{\textbf{Driving Corridor Area (DCA)}}  \\
       \cmidrule(lr){5-9}  \cmidrule(lr){10-14}
        & & & & \textbf{Car} & \textbf{Pedestrian} & \textbf{Cyclist} & \textbf{mAP} & \textbf{mAOS} & \textbf{Car} & \textbf{Pedestrian} & \textbf{Cyclist} & \textbf{mAP} & \textbf{mAOS}  \\
    \hline
      PointPillars (SFPN) & Radar5 & ALBERT \cite{lan2019albert} & HDP & 18.92 & \textbf{\textcolor{red}{9.79}} & 12.47 & 13.73 & 12.91 &  39.20 & \textbf{\textcolor{red}{10.25}} & 14.93 & 21.46 & 20.19   \\
      % SECOND (SFPN) & Radar5 & ALBERT & HDP & 17.70 & 7.67 & 10.58 & 11.98 & 11.06 & 38.17 & 8.62 & 14.95 & 20.58 & 19.77 \\
      CenterFormer & Radar5 & ALBERT & HDP & 17.26 & 6.79 & 9.27 & 11.11 & 10.79 &  19.56 & 9.13 & 12.03 & 13.57 & 13.02 \\
      CenterPoint (Voxel-SFPN)   & Radar5 & ALBERT & HDP  & 18.98 & 5.30 & 14.96 & 13.08 & 12.20 &  40.53 & 8.57 & 15.66 & 21.59 & 20.25  \\
    \hline
      PointPillars (SFPN) & Radar5 & ALBERT & MHCA & 5.18 & 5.76 & 6.63 & 5.86 & 3.58 & 13.34 & 4.36 & 8.79 & 8.83 & 7.66  \\
      % SECOND (SFPN) & Radar5 & ALBERT & MHCA & 4.98 & 3.69 & 4.24 & 4.30 & 2.24 & 10.67 & 3.50 & 7.03 & 7.07 & 6.57 \\
      CenterFormer & Radar5 & ALBERT & MHCA & 4.53 & 3.48 & 4.00 & 4.00 & 2.03  & 8.77 & 3.52 & 6.69 & 6.33 & 5.92 \\
      CenterPoint (Voxel-SFPN)   & Radar5 & ALBERT & MHCA & 5.21 & 4.57 & 5.13 & 4.97 & 3.12 & 12.70 & 4.07 & 7.70 & 8.16 & 7.51 \\
    \hline
     MSSG \cite{cheng2023language} & Radar5 & GRU \cite{chung2014empirical} & - & 12.53 & 5.08 & 8.47 & 8.69 & 7.03 & 18.93 & 7.88 & 9.40 & 12.07 & 11.67\\
     AFMNet \cite{solgi2024transformer} & Radar5 & GRU & - & 11.98 & 6.87 & 9.16 & 9.34 & 7.72 & 18.62 & 8.21 & 10.06 & 12.30 & 11.79 \\
     MSSG \cite{cheng2023language} & Radar5 & ALBERT & - & 16.03 & 5.86 & 10.57 & 10.82 & 8.96 & 25.79 & 8.69 & 12.55 & 15.68 & 14.12 \\
     AFMNet \cite{solgi2024transformer} & Radar5 & ALBERT & - & 16.31 & 6.80 & 10.35 & 11.15 & 9.46 & 26.82 & 8.71 & 12.45 & 15.99 & 14.18 \\
     EDA \cite{wu2023eda} & Radar5 & RoBERTa \cite{liu2019roberta} & - & 13.23 & 6.60 & 8.63 & 9.49 & 8.93 & 23.55 & 8.80 & 11.95 & 14.77 & 13.07 \\
    \hline
    \textbf{T-RadarNet (Ours)} & Radar5 & ALBERT & \textbf{GGF} & \textbf{\textcolor{red}{24.68}} & 9.71 & \textbf{\textcolor{red}{15.74}} & \textbf{\textcolor{red}{16.71}} & \textbf{\textcolor{red}{14.88}} & \textbf{\textcolor{red}{42.58}} & 10.13 & \textbf{\textcolor{red}{17.82}} & \textbf{\textcolor{red}{23.51}} & \textbf{\textcolor{red}{22.37}}  \\
    \hline
      CenterPoint (Pillar-SFPN)   & LiDAR & ALBERT & HDP & \uline{\textbf{\textcolor{blue}{28.16}}} & 6.21 & 17.46 & 17.28 & 16.03 & 43.43 & 6.87 & 27.18 & 25.83 & 24.93   \\
      CenterPoint (Pillar-SFPN)   & LiDAR & ALBERT & MHCA & 6.56 & 5.04 & 5.33 & 5.64 & 4.86 &  13.60 & 4.52 & 7.32 & 8.48 & 7.89 \\
    \hline
     MSSG \cite{cheng2023language} & LiDAR & GRU &  - & 15.38 & 7.52 & 11.67 & 11.52 & 9.76 & 23.27 & 8.68 & 13.51 & 15.15 & 14.75\\
     AFMNet \cite{solgi2024transformer} & LiDAR & GRU &  - & 16.13 & 7.68 & 12.51 & 12.11 & 9.92 & 24.50 & 9.07 & 13.87 & 15.81 & 15.11 \\
     MSSG \cite{cheng2023language} & LiDAR & ALBERT &  - &  18.19 & 7.66 & 11.63 & 12.49 & 10.91 & 29.61 & 10.98 & 14.66 & 18.42 & 16.23 \\
     AFMNet \cite{solgi2024transformer} & LiDAR & ALBERT &  - & 19.50 & 7.92 & 13.56 & 13.66 & 12.18 & 31.68 & 9.23 & 18.90 & 19.94 & 17.59 \\
     EDA \cite{wu2023eda} & LiDAR & RoBERTa & - & 16.10 & 6.91 & 12.88 & 11.96 & 10.10 & 25.10 & 9.28 & 15.73 & 16.70 & 14.91 \\
    \hline
     \textbf{T-RadarNet (Ours)} & LiDAR & ALBERT & \textbf{GGF} & 24.91 & \uline{\textbf{\textcolor{blue}{12.74}}} & \uline{\textbf{\textcolor{blue}{18.67}}} & \uline{\textbf{\textcolor{blue}{18.77}}} & \uline{\textbf{\textcolor{blue}{17.20}}} & \uline{\textbf{\textcolor{blue}{48.98}}} & \uline{\textbf{\textcolor{blue}{14.69}}} & \uline{\textbf{\textcolor{blue}{27.24}}} & \uline{\textbf{\textcolor{blue}{30.30}}} & \uline{\textbf{\textcolor{blue}{29.89}}} \\
    \hline
    \end{tabular}
\end{table*}

\subsection{Deformable FPN}
To enhance representation of text-conditioned three-stage PC features $\{\text{$F_{R|T}^{S_1}$, $F_{R|T}^{S_2}$, $F_{R|T}^{S_3}$}\}$, we first exert the efficient deformable convolutions \cite{wang2023internimage} on three maps to perform adaptive sparse sampling and modeling of key features. For the current element $r_0$ in the one-stage radar map, the deformable convolution can be formulated as 
%Eq. (\ref{eq:dcn}):
\begin{equation}
    y(r_0) = \sum^G _{g=1} \sum^K _{k=1} w_g m_{g}^{k} x_g (r_0 + r_k + \triangle r_{g}^{k}),
    \label{eq:dcn}
\end{equation}
where $G$ and $K$ denote element aggregation groups and the feature dimension. $w_g$ is the projection weights of the group. $m_{g}^{k}$ is the modulation scalar of the $k$-th sampling point while $\triangle r_{g}^{k}$ is the offset of sampling position $r_k$ in the $g$-th group.

Then $F_{R|T}^{S_2}$ $\in \mathbb{R}^{\frac{H}{2} \times \frac{W}{2} \times 2C}$ and $F_{R|T}^{S_3}$ $\in \mathbb{R}^{\frac{H}{4} \times \frac{W}{4} \times 4C}$ are upsampled to the same size as $F_{R|T}^{S_1}$ by transposed convolution. Finally, three-scale feature maps are concatenated to obtain high-resolution aggregated PC features $F_{\rm Agg} \in \mathbb{R}^{H \times W \times 3C}$ with multi-receptive fields.

\subsection{Training Objectives}
The center-based detection head first produces class-wise heatmaps to predict the center location of the detected objects for each class. Afterward, the following properties of each object are supervised following CenterPoint \cite{yin2021center}: the sub-voxel location refinement, the height-above-ground, the 3D size, and the orientation angle. The proposed T-RadarNet is trained with the following loss
% (reducing the quantization error from voxelization and striding of the backbone network)
 % (predicting the elevation shift)
%
\begin{equation}
    {\cal L}_{\rm total} = {\cal L}_{\rm hm} + \beta \sum\limits_{r \in \boldsymbol{\Lambda} } {\cal L}_{{\rm smooth}-\ell_1}(\widehat{\Delta r^{a}}, \Delta r^{a}),
\end{equation}
where ${\cal L}_{\rm hm}$ is the classification loss supervising the heatmap quality of the center-based detection head using a focal loss \cite{Lin_2017_ICCV}; $\boldsymbol{\Lambda} = \left\{ x, y, z, l, h, w, \theta \right\}$ indicates the smooth-$\ell_1$ loss supervising the regression of the box center (for slight modification based on heatmap peak guidance), dimensions, and orientation; and $\beta$ is the weight to balance the two components of the loss, which is set to 0.25 by default.

\begin{figure}
\begin{center}
\includegraphics[width=0.99\linewidth]{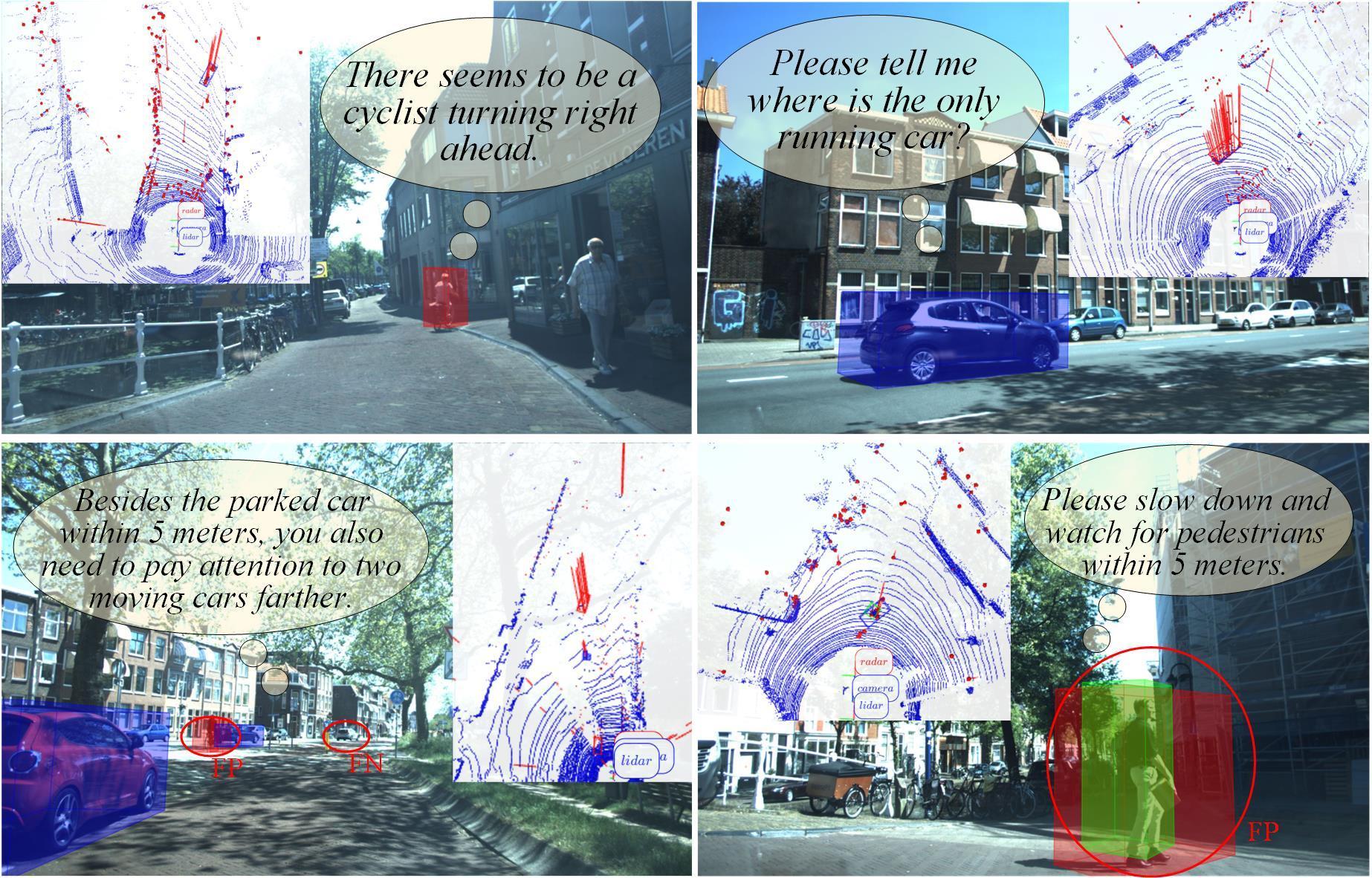}
\end{center}
\vspace{-6mm}
\caption{Prediction by T-RadarNet. The first row presents correct cases while the second shows problematic cases (\textcolor{red}{FN}: False Negative, \textcolor{red}{FP}: False Positive).}
\label{fig:pred_cases}
\end{figure}
% \vspace{-2mm}

\section{Experiments}
\subsection{Implementation Settings}

\textbf{Model Settings:} For T-RadarNet, we set the token length as 30 upon ALBERT \cite{lan2019albert} and other text encoders for comparison. $N$ in the pillar encoder is set as 10 and 32 for radar and LiDAR, respectively, while $C$ is 64. For comparison, we \textbf{firstly} select PC-based detectors with various paradigms including PointPillars (pillar-based) \cite{lang2019pointpillars}, CenterPoint (voxel and anchor-free) \cite{yin2021center}, and CenterFormer (transformer-based) \cite{zhou2022centerformer}. \textbf{Secondly}, fusion methods including inductive bias-based HDP \cite{zhu2022seqtr}, attention-based MHCA \cite{wu2023referring} are implemented to compare with our GGF. \textbf{Thirdly}, different necks, \ie, SecondFPN \cite{yan2018second}, ASPP \cite{li2023pillarnext} and CSP-FPN \cite{ge2021yolox} are included for comparison with proposed Deformable FPN. \textbf{Lastly}, we also compare proposed T-RadarNet with dedicated SOTA 3D PC grounding model MSSG \cite{cheng2023language}, AFMNet \cite{solgi2024transformer} and EDA \cite{wu2023eda}. 

\textbf{Dataset Settings:} We train and test models on Talk2Radar dataset with three type of objects, \ie, Car, Cyclist, and Pedestrian, using both 4D radar and LiDAR. Moreover, to validate the generalization of T-RadarNet, we also train and test it on Talk2Car dataset \cite{deruyttere2019talk2car}, which was built upon nuScenes \cite{caesar2020nuscenes} by providing textual prompt for LiDAR PC.
%, making each sample contains one textual prompt for LiDAR 3D grounding.
 %we also train and test it on Talk2Car \cite{deruyttere2019talk2car} built on nuScenes \cite{caesar2020nuscenes}, which provides LiDAR PC and each sample contains one textual prompt for a unique object.

\textbf{Training and Evaluation Settings:} For Talk2Radar, all models are trained on four RTX A4000 with a batch size of 4 for 80 epochs. The initial learning rate is set 1e-3 with a cosine scheduler. We choose AdamW for optimization with a weight decay of 5e-4. For Talk2Car, we set the initial learning rate as 1e-2 while other settings are kept the same with Talk2Radar.

\subsection{Quantitative Results}

\textbf{Overall Performances:} As Table \ref{tab:benchmark_compare} shows, on the whole, T-RadarNet outperforms other models in most aspects whatever on 4D radar or LiDAR. For other models, we find that the models based on pillar encoding perform better than those based on voxels, and the models based on self-attention do not significantly outperform those based on convolution. Regarding fusion, models using MHCA perform worse than those using gating and dot-product operations. This phenomenon is particularly evident with five-frame radar data (Radar5). \textbf{Performances on Various Prompts:} As Table \ref{tab:prompt_compare} shows, when we compare the performance of T-RadarNet on prompts with queries of types on motion, depth and velocity, the radar-based model outperforms LiDAR-based on prompts regarding motion and velocity, which is more obvious on velocity-based prompts. Besides, as LiDAR can reason the object motion upon PC object appearances, the gap between radar and LiDAR on motion query is not large. For the depth-related prompts, LiDAR obtains the advantage to some extent. \textbf{Precision by Piece-wise Depths:} Table \ref{tab:distance_compare} presents the mAP of predicted objects by segmented depths upon 5-frame radar. Small-object grounding is doubtless a challenge. \textbf{Generalization Performances:} Table \ref{tab:talk2car_compare} shows the generalization evaluation results of T-RadarNet on Talk2Car for 3D REC, our T-RadarNet outperforms the baseline and the other three models, which proves the effectiveness and generalization of T-RadarNet for PC-based 3D REC.

\begin{table}
\setlength\tabcolsep{2.0pt}
    \caption{Comparison (\textbf{mAP}) of radar and LiDAR upon various prompts}
    \vspace{-3mm}
    \label{tab:prompt_compare}
    \centering
    \begin{tabular}{c|ccc|ccc|ccc}
    \hline
      \textbf{Prompt} & \multicolumn{3}{c}{\textbf{Motion}} & \multicolumn{3}{c}{\textbf{Depth}} & \multicolumn{3}{c}{\textbf{Velocity}}  \\
    \hline
       \textbf{Sensors}  & \textbf{Car} & \textbf{Pedes} & \textbf{Cyclist} & \textbf{Car} & \textbf{Pedes} & \textbf{Cyclist} & \textbf{Car} & \textbf{Pedes} & \textbf{Cyclist}  \\
    \hline
     Radar1 & 26.10 & 5.33 & 11.07 & 32.68 & 12.69 & 24.72 & 26.92 & 13.99 & 18.45  \\
     Radar3 & 35.92 & 11.05 & 15.43 & 40.68 & 18.66 & 31.69 & 33.78 & 20.51 & \textbf{25.78}\\
     Radar5 & \textbf{36.72} & \textbf{11.54} & \textbf{16.27} & 42.50 & 19.63 & 32.03 & \textbf{35.50} & \textbf{20.79} & 25.73\\
     LiDAR & 33.56 & 8.80 & 13.58 & \textbf{45.68} & \textbf{20.54} & \textbf{38.60} & 12.63 & 7.63 & 8.24\\
    \hline     
    \end{tabular}  
\end{table}

\begin{table}
\setlength\tabcolsep{6.7pt}
    \caption{Statistics of predicted objects \textbf{mAP} by depth upon 5-frame radar}
    \vspace{-3mm}
    \label{tab:distance_compare}
    \centering
    \begin{tabular}{c|cccccc}
    \hline
      \textbf{Objects} & \textbf{0-10 (m)} & \textbf{10-20} & \textbf{20-30} & \textbf{30-40} & \textbf{40-50} & \textbf{50+}  \\
    \hline
       \textbf{Car} & 42.57 & 44.57 & 23.26 & 18.36 & 10.04 & 1.32 \\
       \textbf{Pedestrian} & 15.54 & 7.17 & 3.14 & 2.06 & 4.54 & 0.0  \\
       \textbf{Cyclist} & 29.15 & 11.39 & 14.14 & 3.94 & 1.78 & 0.0  \\
    \hline
    \end{tabular}
\end{table}

\begin{table}
\setlength\tabcolsep{14.5pt}
    \caption{Performances on Talk2Car benchmark for LiDAR-based 3D REC, where \textbf{AP$_\text{A}$} and \textbf{AP$_\text{B}$} follow MSSG \cite{cheng2023language} that define different IoU thresholds.}
    \vspace{-3mm}
    \label{tab:talk2car_compare}
    \centering
    \begin{tabular}{c|cccc}
    \hline
    \multirow{2}[2]{*}{\textbf{Models}} & \multicolumn{2}{c}{\textbf{BEV AP}} & \multicolumn{2}{c}{\textbf{3D AP}}  \\
    \cmidrule{2-3} \cmidrule{4-5} 
         & \textbf{AP$_\text{A}$} & \textbf{AP$_\text{B}$} & \textbf{AP$_\text{A}$} & \textbf{AP$_\text{B}$} \\
    \hline
    Talk2Car \cite{deruyttere2019talk2car} & 30.6 & 24.4 & 27.9 & 19.1 \\
    MSSG \cite{cheng2023language} & 27.8 & 26.1 & 31.9 & 20.3\\
    EDA \cite{wu2023eda} & 37.0 & 29.8 & 37.2 & 20.4 \\
    AFMNet \cite{solgi2024transformer} & 45.3 & 33.1 & 41.9 & 20.7 \\
    \hline
    T-RadarNet & \textbf{52.8} & \textbf{39.9} & \textbf{47.2} & \textbf{30.5} \\
    \hline
    \end{tabular}
\end{table}

\begin{table}
\setlength\tabcolsep{5.4pt}
    \caption{Ablation comparison of T-RadarNet upon 5-Frame radar data}
    \vspace{-3mm}
    \label{tab:ablation_exp}
    \centering
    \begin{tabular}{c|ccc|ccc}
    \hline
    \multirow{2}[2]{*}{\textbf{Methods}}  & \multicolumn{3}{c}{\textbf{EAA}} & \multicolumn{3}{c}{\textbf{DCA}} \\
     \cmidrule(lr){2-4} \cmidrule(lr){5-7}
     & \textbf{Car} & \textbf{Ped} & \textbf{Cyc} & \textbf{Car} & \textbf{Ped} & \textbf{Cyc} \\
    \hline
      \multicolumn{7}{c}{\textbf{Gated Graph Fusion}} \\
    \hline
      \textbf{GConv} $\rightarrow$ Conv  & 20.03 & 7.56 & 13.70 & 36.56 & 7.99 & 13.58 \\
      \textbf{MaxPool} $\rightarrow$ AvgPool & 21.40 & 8.04 & 13.88 & 38.67 & 7.87 & 13.78 \\
      \textbf{GConv + MaxPool} & \textbf{24.68} & \textbf{9.71} & \textbf{15.74} & \textbf{42.58} & \textbf{10.13} & \textbf{17.82} \\
    \hline
      \multicolumn{7}{c}{\textbf{Fusion Methods}} \\
    \hline
    HDP \cite{zhu2022seqtr} & 20.07 & 6.52 & 14.86 & 39.53 & 8.62 & 15.85\\
    MHCA \cite{wu2023referring} & 7.79 & 5.02 & 5.36 & 13.69 & 5.81 & 9.35 \\
    % MHLCA & 6.97 & 5.20 & 5.05 & 12.54 & 6.03 & 8.96 \\
    \textbf{GGF} & \textbf{24.68} & \textbf{9.71} & \textbf{15.74} & \textbf{42.58} & \textbf{10.13} & \textbf{17.82} \\
    \hline
      \multicolumn{7}{c}{\textbf{Necks (Feature Pyramid Networks)}} \\
    \hline
    ASPP \cite{li2023pillarnext} & 23.01 & 9.56 & 14.59 & 38.06 & 9.27 & 16.60 \\
    SecondFPN \cite{yan2018second} & 23.57 & 9.26 & 14.03 & 41.56 & 9.02 & 16.42\\
    CSP-FPN \cite{ge2021yolox} & 23.53 & 9.70 & 14.38 & 41.58 & 9.31 & 17.03\\
    \textbf{DeformableFPN} & \textbf{24.68} & \textbf{9.71} & \textbf{15.74} & \textbf{42.58} & \textbf{10.13} & \textbf{17.82} \\
    \hline
      \multicolumn{7}{c}{\textbf{Text Encoders}} \\
    \hline
    Bi-GRU \cite{chung2014empirical} & 18.74 & 7.57 & 10.89 & 24.60 & 7.98 & 11.50 \\
    RoBERTa \cite{liu2019roberta} & 24.22 & 9.53 & \textbf{16.12} & \textbf{42.62} & 10.03 & 16.66  \\
    \textbf{ALBERT} & \textbf{24.68} & \textbf{9.71} & 15.74 & 42.58 & \textbf{10.13} & \textbf{17.82} \\
    \hline
    \end{tabular}
\end{table}

\subsection{Ablation Experiments}
Table \ref{tab:ablation_exp} presents the ablation results of T-RadarNet. \textbf{GGF:} When we replace the MRConv4D (GConv) with normal convolution, mAP drops remarkably, which is the same circumstance with pooling for textual semantics abstraction. \textbf{Fusion Methods:} Our proposed GGF presents a better result than other fusion methods, implying a more efficient approach to aligning and embedding textual semantics to PCs. Although MHCA can model global cross-modal similarity, its performance is sub-optimal when dealing with irregular PCs containing both objects and ghosting in radar data. \textbf{Necks:} Deformable FPN outperforms all other necks, indicating its preeminent capability of representing PCs. \textbf{Text Encoders:} Pretrained transformer-based text encoders outperform RNN-based text encoders remarkably, indicating the significance of attention weighting on text encoding for the following fusion. 

% \textbf{Fusion Locations.} Previous works \cite{roh2022languagerefer,huang2022multi,guo2023viewrefer,chen2022language,solgi2024transformer} prove the effectiveness of PC-text fusion on the location of models, including neck-to-head and backbone-to-neck. In Talk2Radar, we find that fusing two modalities on the output of the backbone is better than doing that after the neck.

\subsection{Visualization and Discussion}
Fig. \ref{fig:pred_cases} presents the prediction results by T-RadarNet based on the 5-frame accumulated radar PC, including visualization on the image plane and BEV view in the PC context. For the correct prediction results in the first row, we can see our T-RadarNet can understand textual prompts and localize the correct objects among radar PC contexts, including single and multiple referent objects. However, there are still some problems with this task, in the second row, we can see the serious false positive cases in the predicted bounding boxes, implying that efficient representation of radar PC features and adaptive filtering of non-object clutter are still a challenge.

\section{Conclusion}
This paper proposes a novel task: 4D mmWave radar-based 3D referring expression comprehension (3D visual grounding), aimed at enhancing embodied intelligence and interactive perception with an all-weather, low-cost sensor. We introduce the first dataset for this task, Talk2Radar, which fully leverages the object detection capabilities of 4D mmWave radar. Talk2Radar includes a rich collection of textual prompts and object distributions, alongside LiDAR data. Through extensive experiments, we establish a corresponding benchmark by proposing an efficient radar-based 3D REC model, T-RadarNet. Within T-RadarNet, we design an effective module, Gated Graph Fusion, for the alignment and fusion of textual and 4D radar point cloud features. Additionally, we propose Deformable FPN to adequately model irregular and sparse point cloud features. With these contributions, we aim to advance the interactive perception capabilities of 4D radar for environmental understanding in autonomous driving.
% This paper proposes a novel task: 4D mmWave radar-based 3D referring expression comprehension (3D visual grounding), aimed at embodied intelligence and interactive perception with full weather operational, low-cost sensor. The first dataset on such task, Talk2Radar, has been presented by fully exploiting object detection characteristics of 4D mmWave radar. Talk2Radar includes abundant textual prompts and object distributions, as well as LiDAR. Through extensive experiments, the corresponding benchmark has been provided by proposing an efficient radar-based 3D REC model, T-RadarNet. In T-RadarNet, we design an effective module, Gated Graph Fusion, for the alignment and fusion of textual and 4D radar point cloud features. Additionally, Deformable FPN is proposed to adequately model irregular and sparse point cloud features. Having all the aforementioned contributions, we aim to advance the interactive perception capabilities of 4D radar for environmental understanding in autonomous driving.

% \begin{table}[htbp]
% \caption{Table Type Styles}
% \begin{center}
% \begin{tabular}{|c|c|c|c|}
% \hline
% \textbf{Table}&\multicolumn{3}{|c|}{\textbf{Table Column Head}} \\
% \cline{2-4} 
% \textbf{Head} & \textbf{\textit{Table column subhead}}& \textbf{\textit{Subhead}}& \textbf{\textit{Subhead}} \\
% \hline
% copy& More table copy$^{\mathrm{a}}$& &  \\
% \hline
% \multicolumn{4}{l}{$^{\mathrm{a}}$Sample of a Table footnote.}
% \end{tabular}
% \label{tab1}
% \end{center}
% \end{table}

\normalem
%\small
\footnotesize
\bibliographystyle{IEEEbib}
\bibliography{ICRA2025_v3}

\end{document}